# EDGE DETECTION IN RADAR IMAGES USING WEIBULL DISTRIBUTION

Ali El-Zaart , Wafaa Kamel Al-Jibory

Department of Mathematics and Computer Science

Faculty of Science, Beirut Arab University

Beirut –Lebanon

dr_elzaart@yahoo.com, waf_jibory@yahoo.com

*ABSTRACT*

*Radar images can reveal information about the shape of the surface terrain as well as its physical and biophysical properties. Radar images have long been used in geological studies to map structural features that are revealed by the shape of the landscape. Radar imagery also has applications in vegetation and crop type mapping, landscape ecology, hydrology, and volcanology. Image processing is using for detecting for objects in radar images. Edge detection; which is a method of determining the discontinuities in gray level images; is a very important initial step in Image processing. Many classical edge detectors have been developed over time. Some of the well-known edge detection operators based on the first derivative of the image are Roberts, Prewitt, Sobel which is traditionally implemented by convolving the image with masks. Also Gaussian distribution has been used to build masks for the first and second derivative. However, this distribution has limit to only symmetric shape. This paper will use to construct the masks, the Weibull distribution which was more general than Gaussian because it has symmetric and asymmetric shape. The constructed masks are applied to images and we obtained good results.*

*KEYWORDS*

*Edge detection; Image processing; Weibull Distribution; Gradient; Radar Image.*

## I- INTRODUCTION

Unlike most remote sensing systems, which rely on the sun as an energy source, imaging radar systems are active sensors that "illuminate" surface features with broadcast microwave energy and record a returned signal. Most imaging radar systems produce microwaves with wavelengths between 1 cm and 1 meter, longer than the wavelengths used in weather radar to detect rain and snow. As a result imaging radar systems can be used to map surface features day or night and in almost any weather conditions [2]. An edge is usually a step change in intensity in an image (radar image). It corresponds to the boundary between two regions or a set of points in the image where luminous intensity changes very sharply[10]. The decision of whether pixel is an edge point or not bases on how much its local neighbours respond to a certain edge detector [13]. Over the years, many methods have been proposed for detecting edges in images. Some of the earlier methods, such as the Sobel and Prewitt detectors [11], used local gradient operators [12] to obtain spatial filter masks. The procedure is to compute the sum of products of the mask coefficients with the intensity values in the region encompassed by the mask [3]. Also the Canny edge detector which depends on the Gaussian distribution for obtaining the operators for the gradient and Laplacian masks is a well-known edge detector [9]. In this paper we propose method that will use Weibull Distribution instead of Gaussian distribution to obtaining edge detection operators. The advantage of this method is that Gaussian distribution has limitation to only symmetric shape but Weibull Distribution has symmetric and asymmetric shape.

The rest of this paper is organized as follows. Section 2 introduces Gradient Edge Detection. Section 3 explains the Most famous Edge Detector. Gradient of Weibull Edge Detector is displayed in Section 4. Experimental result is shown in Section 5. Finally this paper presents conclusion and future work in Section 6.

## II- GRADIENT EDGE DETECTION

Gradient is a vector which has certain magnitude and direction. In image processing, the gradient is the change in gray level with direction. This can be calculated by taking the difference in value of neighbouring pixels

Where $\nabla f$ is first order derivative of $f(x,y)$ define as:

$$\nabla^2 f(x, y) = \frac{\partial^2 f(x, y)}{\partial x^2} + \frac{\partial^2 f(x, y)}{\partial y^2} \qquad (1)$$

The magnitude of this vector, denoted magn(f), Where

$$\text{magn}(\nabla f) = \sqrt{\left(\frac{\partial f}{\partial x}\right)^2 + \left(\frac{\partial f}{\partial y}\right)^2} = \sqrt{G_x^2 + G_y^2} \qquad (2)$$

The direction of the gradient vector, denoted dir(f), Where

$$\text{dir}(\nabla f) = \tan^{-1}(G_y/G_x) \qquad (3)$$

The magnitude of gradient provides information about the strength of the edge and the direction of gradient is always perpendicular to the direction of the edge.

## III- THE FAMOUSE EDGE DETECTOR

Before introducing the proposed algorithm, this section reviews some of the main edge detection methods, such as the Sobel method and Gradient of Gaussian edge detector.

### 3.1. The Sobel Edge Detector

The Sobel method [3] utilizes two masks, $S_x$ and $S_y$, shown in fig. 1, to do convolution on the gray image and then obtain the edge intensities $G_x$ and $G_y$ in the vertical and horizontal directions, respectively. The edge intensity of the mask center is defined as $|G_x|+|G_y|$. If the edge intensity of each pixel is larger than an appropriate threshold T, then the pixel is regarded as an edge point. Unfortunately, the edge line detected by Sobel method is usually thicker than the actual edge [16].

|    |    |    |   |    |    |    |
|----|----|----|---|----|----|----|
| -1 | 0  | 1  |   | -1 | -2 | -1 |
| -2 | 0  | 2  |   | 0  | 0  | 0  |
| -1 | 0  | 1  |   | 1  | 2  | 1  |

        Sx                  Sy

Figure 1. Two convolution masks in Sobel method.

### 3.2. The Gradient of Gaussian edge detector

An edge detection operator can reduce noise by smoothing the image, but this adds uncertainly to the location of the edge: or the operator can have greater sensitivity to the presence of edges, but this will increase the sensitivity of the operator to noise. The type of liner operator that provides the best compromise between noise immunity and localization, while retaining the advantages of Gaussian filtering is the first derivative of a Gaussian. This operator corresponds to smoothing an image with a Gaussian function and then computing the gradient. The gradient can be numerically approximated by using the standard finite-difference approximation for the first partial derivative in the x and y directions. The operator that is the combination of a Gaussian smoothing filter and a gradient approximation is not rotationally symmetric. The operator is symmetric along the edge and antisymmetric perpendicular to the edge (along the line of the gradient). This means that the operator is sensitive to the edge in the direction of steepest change, but is insensitive to the edge and acts as a smoothing operator in the direction along the line

## IV- GRADIENT OF WEIBULL EDGE DETECTOR

The Gaussian distribution is the most popularly used distribution model in the field of pattern recognition. It is used to build masks for the first and second derivative. However, it has limit to only symmetric shape. We will propose new method that uses Weibull Distribution which is more general than Gaussian because it has symmetric and asymmetric shape.

In this section we will explain the characteristic of 1D Weibull distribution and how calculate 2D Weibull distribution from 1D. The 1D Weibull distributions have the probability density function is given by:

$$1DW(x;\alpha,\beta) = \begin{cases} \alpha\beta x^{\beta-2} e^{-\alpha x^\beta}(\beta - 1 - \alpha\beta x^\beta) & x > 0 \\ 0 & \text{elsewhere} \end{cases} \quad (4)$$

The distribution is skewed with a longer tail to the right of the mean as shown in Fig .3. Where $\mu$ (mean) is scale parameter and $\sigma$ (standard deviation) shape parameters and $x > 0$. For the same $\sigma$, the function skewness increases as $\mu$ increases and for the same $\mu$, the function is rises very sharply in the beginning when the value of $\sigma$ significantly greater than 1[8].

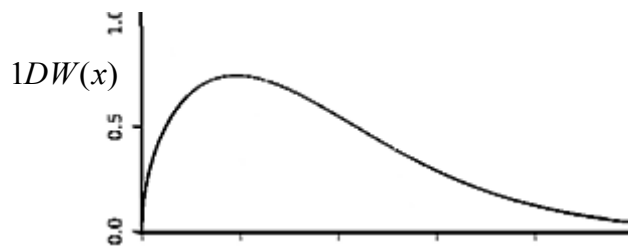

Figure 3. Probability density function of the Weibull distribution

We can calculate the 2D Weibull distribution by produce of: $1DW(x;\mu,\sigma)$ and $1DW(y;\mu,\sigma)$ where $f$ is "Eq. (4)" given by:

$$2DW(x,y;\mu,\sigma) = 1DW(x;\mu,\sigma) \times 1DW(y;\mu,\sigma) = \begin{cases} \alpha^2 \beta^2 x^{\beta-1} y^{\beta-1} e^{-\alpha(x^\beta + y^\beta)} & x>0, y=0 \\ 0 & \text{Elsewhere} \end{cases} \quad (5)$$

In the following section we will explain how perform smoothing using 2D Weibull Distribution then we will explain the applying of edge detection using the gradient of the 2D Weibull Distribution. After that we will present the edge detection using the Laplacian of 2D Weibull Distribution.

### 4.1. Smoothing Using 2D Weibull Distribution

We can construct smoothing filter from 2D Weibull Distribution. The method is to create general 3x3 mask from 2D Weibull Distribution as show bellow:

| $2DW(x-d, y-d)$ | $2DW(x-d, y)$ | $2DW(x-d, y+d)$ |
|---|---|---|
| $2DW(x, y-d)$ | $2DW(x, y)$ | $2DW(x, y+d)$ |
| $2DW(x+d, y-d)$ | $2DW(x+d, y)$ | $2DW(x+d, y+d)$ |

For 5x5 mask the left corner would be $2DW(x-2d, y-2d)$ and for 7x7 would be $2DW(x-3d, y-3d)$ also for 9x9 $2DW(x-4d, y-4d)$ and so on. The sum of all values of the mask must be 1 because it is smoothing mask. The value d is incremental and the best value of incremental we can determine through the experiment.

### 4.2. Edge Detection using the Gradient of the 2D Weibull Distribution

We can construct the gradient mask of $2DW(x,y)$ by obtaining the first partial derivative of $x, y$ for $2DW(x,y)$. The first x derivative for $2DW(x,y)$ is given by:

$$M_x 2DW(x,y) = \frac{\partial}{\partial x} 2DW(x,y) \begin{cases} \alpha^2 \beta^2 x^{\beta-2} y^{\beta-1} e^{-\alpha(x^\beta+y^\beta)}(\beta-1-\alpha\beta x^\beta) & x>0, y=0 \\ 0 & \text{Elsewhere} \end{cases} \quad (6)$$

The first y derivative for $2DW(x,y)$ is given by:

$$= \begin{cases} \alpha^2 \beta^2 y^{\beta-2} x^{\beta-1} e^{-\alpha(x^\beta+y^\beta)}(\beta-1-\alpha\beta y^\beta) & x>0, y=0 \\ 0 & \text{Elsewhere} \end{cases} \quad (7)$$

Using the first x derivative for $2DW(x,y)$ we can construct $M_x$ mask:

| $M_x 2DW$ $(x-incx, y-incy)$ | $M_x 2DW$ $(x-incx, y)$ | $M_x 2DW$ $(x-incx, y+incy)$ |
|---|---|---|
| $M_x 2DW$ $(x, y-incy)$ | $M_x 2DW(x, y)$ | $M_x 2DW$ $(x, y+incy)$ |
| $M_x 2DW$ $(x+incx, y-incy)$ | $M_x 2DW$ $(x+incx, y)$ | $M_x 2DW$ $(x+incx, y+incy)$ |

Using the first y derivative for $2DW(x, y)$ we can construct $M_y$ mask:

| $M_y 2DW$ $(x-incx, y-incy)$ | $M_y 2DW$ $(x-incx, y)$ | $M_y 2DW$ $(x-incx, y+incy)$ |
|---|---|---|
| $M_y 2DW$ $(x, y-incy)$ | $M_y 2DW(x, y)$ | $M_y 2DW$ $(x, y+incy)$ |
| $M_y 2DW$ $(x+incx, y-incy)$ | $M_y 2DW$ $(x+incx, y)$ | $M_y 2DW$ $(x+incx, y+incy)$ |

We need to calculate two increments one for x and one for y.

The sum of the gradient mask should be zero. So after constructing the masks, they should be normalized. The positive values are added and then divided by their sum to obtain 1 and the negative values are computed in the same way to obtain –1.

The obtained masks at alpha = 1, beta =2 are as follows:

$M_x =$

| 0.6951 | 1.5025 | 0.9850 |
|---|---|---|
| 0 | 0 | 0 |
| -0.3538 | -0.7648 | -0.5014 |

$M_y =$

| 0.6951 | 0 | -0.3538 |
|---|---|---|
| 1.5025 | 0 | -0.7648 |
| 0.9850 | 0 | -0.5014 |

After normalization we got these results:

$M_x =$

| 0.2184 | 0.4721 | 0.3095 |
|---|---|---|
| 0 | 0 | 0 |
| -0.2184 | -0.4721 | -0.3095 |

$M_y =$

| 0.2184 | 0 | -0.2184 |
|---|---|---|
| 0.4721 | 0 | -0.4721 |
| 0.3095 | 0 | -0.3095 |

The obtained masks at alpha = 1, beta =3 are as follows:

$M_x =$

| 0.1550 | 1.2799 | 0.9149 |
|---|---|---|
| 0.1785 | 1.4738 | 1.0535 |
| -0.2606 | -2.1526 | -1.5388 |

$M_y =$

| 0.1550 | 0.1785 | -0.2606 |
|---|---|---|
| 1.2799 | 1.4738 | -2.1526 |
| 0.9149 | 1.0535 | -1.5388 |

After normalization we got these results:

$$M_x = \begin{bmatrix} 0.0307 & 0.2532 & 0.1810 \\ 0.0353 & 0.2915 & 0.2084 \\ -0.0660 & -0.5447 & -0.3894 \end{bmatrix} \quad M_y = \begin{bmatrix} 0.0307 & 0.0353 & -0.0660 \\ 0.2532 & 0.2915 & -0.5447 \\ 0.1810 & 0.2084 & -0.3894 \end{bmatrix}$$

## V- EXPERIMENTAL RESULTS

We present in this section our experimental results of using Weibull Distribution in detecting edges using Gradient of this distribution and compare this result with Sobel.

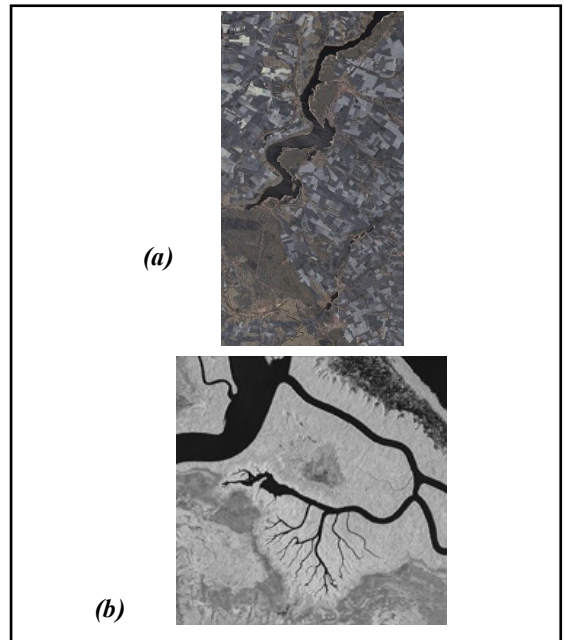

Figure 4. Original Images used in edge detection

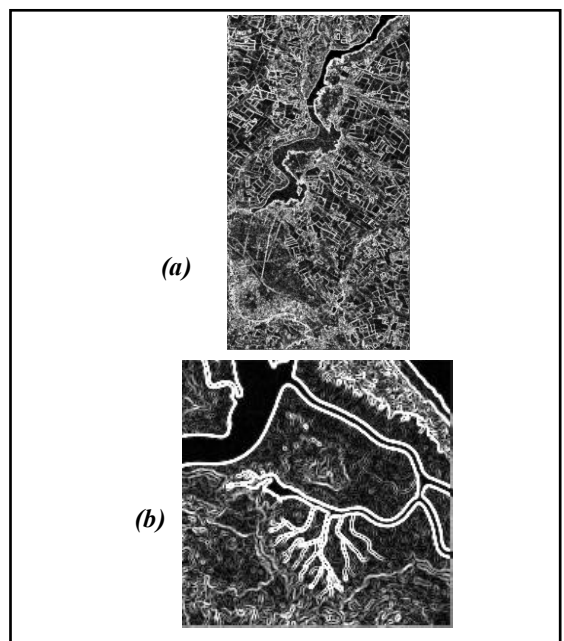

Figure 5. Sobel Results

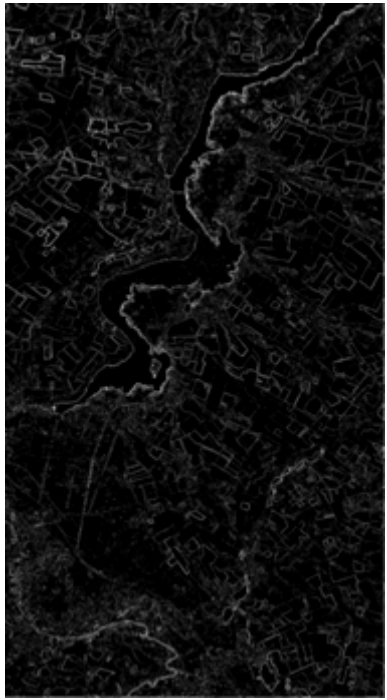
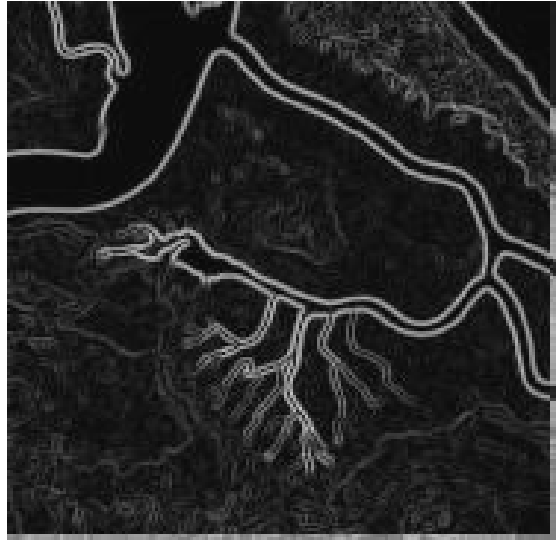

Alpha=1,Beta=2

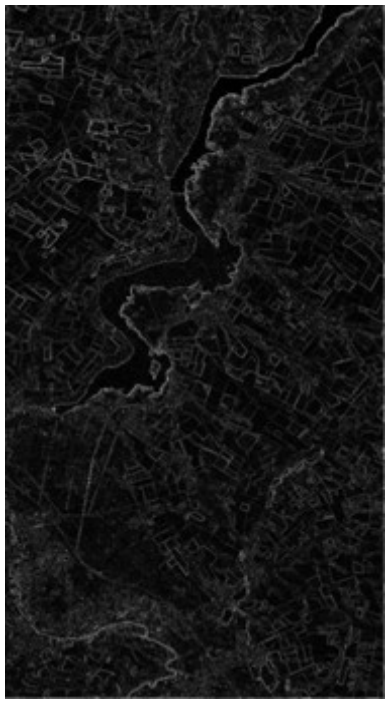
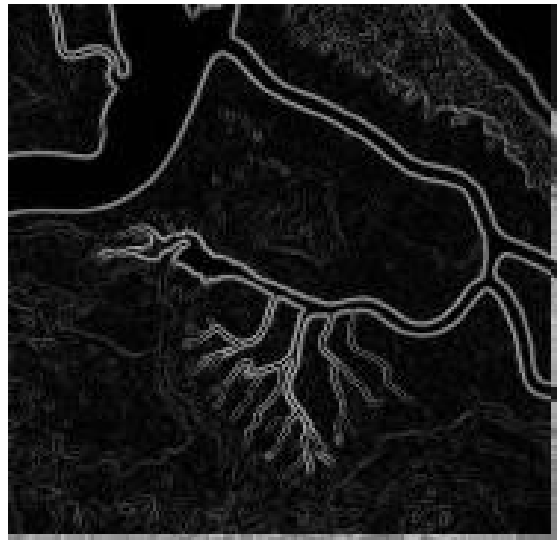

Alpha=1,Beta=3

Figure 6. Results of the new Gradient detector of weibull of size 3x3

From Fig.5 and Fig.6 we can notice that the Weibull distribution is better than the gradient of Gaussian method it produces thinner edges and is less sensitive to noise. This is because the Gaussian distribution has limit to only symmetric shape but Weibull distribution has symmetric and asymmetric shape.

## VI- CONCLUSION AND FUTURE WORK

Smoothing is a prior step in any edge detection process to suppress as much noise as possible. Edge detection using first derivative depend on Gaussian works well when the image contains sharp intensity transitions and low noise, while Edge detection using LOG make better localization, especially when the edges are not very sharp. We proposed new method that uses Weibull Distribution instead of Gaussian distribution to build masks for the first and second derivative and for smoothing image.

## VII- ACKNOLEDGMENT

We thank the Dr. Toufic El Arwadi for his help on the mathematical calculation

## REFERENCES

[1] J.K.Kim, J.M.Park, K.S.Song and H.W.Park " Adaptive Mammographic Image Enhancement Using First Derivative and Local Statistics" IEEE Transactions on medical imaging, Vol.16,No.5,October 1997.

[2] Randall B. Smith. "Interpreting Digital Radar Images". MicroImages, January 2012.

[3] R. Gonzalez and R.Woods, "Digital image processing," 3rd Edition, Prentice Hall, New York, 2008, pp. 695.

[4] E.Trucco and A.Verri." Introductory Techniques for 3-D Computer Vision" Prentice Hall, New York ,1998. Chapter 4.2.

[5] C. Kanga,Wen-JuneWang, "Anovel edge detection method based on the maximizing objective function," Taiwan, April 2007.

[6] J.Canny. "A Computational Approach to Edge Detection Edge Detection".IEEE Transactions on Pattern Analysis and machine intelligence, Vol. PAMI-8, NO. 6, November 1986.

[7] N.T. Thomopoulos, Arvid C. Johnson." Tables And Characteristics of the Standardized Lognormal Distribution", Proceedings of the Decision Sciences Institute, 2003, pp. 1031-1036.

[8] Relia Soft." Characteristics of the Lognormal Distribution".2006.

[9] E. Al-Owaisheque, A. Al-Owaisheq and A. El-Zaart, "A New Edge Detector Using 2D Beta Distribution". Proceedings of the 3rd IEEE International Conference on Information & Communication Technologies : from Theory to Application. April, 9-11, 2008, Syria.

[10] N.Nain, G. Jindal, A. Garg and A. Jain. "Dynamic Thresholding Based Edge Detection". Proceedings of the World Congress on Engineering 2008 Vol I. WCE 2008, July 2 - 4, 2008, London, U.K.

[11] R. J. Qian and T. S. Huang, "Optimal edge detection in two-dimensional images," Proc. Image Understanding Workshop, 1994, pp. 1581–1588.

[12] M. Basu, "Gaussian-Based Edge-Detection Methods A Survey," IEEE Transactions on systems, man, and cybernetics part C: applications and reviews, vol. 32, no. 3, August 2002.

[13] R. Gurcan, Isin Erer and Sedef Kent, "An Edge Detection Method Using 2-D Autoregressive Lattice Prediction Filters for Remotely Sensed Images," Istanbul Technical University Maslak, İstanbul, Turkey 2004.

[14] H.Chidiac et Djemel Ziou, "Classification of Image Edges," Universit´e de Sherbrooke, Sherbrooke (Qc), Canada, J1K 2R1.

[15] B. G. Schunck, "Edge detection with Gaussian filters at multiple scales," in Proc. IEEE Comp. Soc. Work. Comp. Vis., 1987.

[16] L.R. Liang, C.G. Looney, Competitive fuzzy edge detection, Appl. Soft Comput. J. 3 (2003) 123–137.

**Ali El-Zaart** was a senior software developer at Department of Research and Development, Semiconductor Insight, Ottawa, Canada during 2000-2001. From 2001 to 2004, he was an assistant professor at the Department of Biomedical Technology, College of Applied Medical Sciences, King Saud University. From 2004-2010 he was an assistant professor at the Department of Computer Science, College of computer and information Sciences, King Saud University. In 2010, he promoted to associate professor at the same department. Currently, his is an associate professor at the department of Mathematics and Computer Science, Faculty of Sciences; Beirut Arab University. He has published numerous articles and proceedings in the areas of image processing, remote sensing, and computer vision. He received a B.Sc. in computer science from the Lebanese University; Beirut, Lebanon in 1990, M.Sc. degree in computer science from the University of Sherbrooke, Sherbrooke, Canada in 1996, and Ph.D. degree in computer science from the University of Sherbrooke, Sherbrooke, Canada in 2001. His research interests include image processing, pattern recognition, remote sensing, and computer vision.

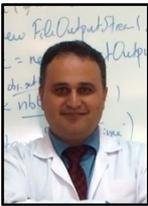

**Wafaa Kamil S. Al-jibory:** Currently he is a master student in Beirut Arab University, Department of Mathematics and Computer Science, Beirut, Lebanon (BAU).

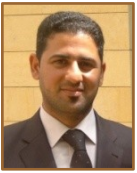